\documentclass[twocolumn]{article}

\usepackage[utf8]{inputenc}
\usepackage{tipa}

\usepackage[a4paper, left=1.8cm, right=1.8cm, top=2.3cm, bottom=2.4cm]{geometry}
\usepackage{fancyhdr}
\usepackage{amsmath}
\usepackage{amsfonts}
\usepackage{amssymb}
\usepackage{verbatim}
\usepackage{graphicx}
\graphicspath{{media/}}
\usepackage{float}
\usepackage{booktabs}   
\usepackage{longtable}
\usepackage{tabularx}
\usepackage[table,xcdraw]{xcolor}
\usepackage{tikz}
\usetikzlibrary{calc, positioning, fit, arrows, shapes.geometric}
\usepackage{hyperref}
\usepackage[numbers]{natbib}
\usepackage{tablefootnote}
\usepackage{enumitem}
\usepackage[normalem]{ulem}
\useunder{\uline}{\ul}{}
\usepackage{caption}
\usepackage{subcaption}
\usepackage{svg}
\usepackage{float}
\usepackage{booktabs}
\usepackage{cuted}

\begin{document}

\title{TGraphX: Tensor-Aware Graph Neural Network\\ for Multi-Dimensional Feature Learning}

\author{Arash Sajjadi$^{*}$ \\ University of Saskatchewan \\ \texttt{arash.sajjadi@usask.ca} \\ \and Mark Eramian$^{*}$ \\ University of Saskatchewan \\ \texttt{eramian@cs.usask.ca}}

\date{\today}

\maketitle

\begin{abstract}
TGraphX presents a novel paradigm in deep learning by unifying convolutional neural networks (CNNs) with graph neural networks (GNNs) to enhance visual reasoning tasks. Traditional CNNs excel at extracting rich spatial features from images but lack the inherent capability to model inter-object relationships. Conversely, conventional GNNs typically rely on flattened node features, thereby discarding vital spatial details. TGraphX overcomes these limitations by employing CNNs to generate multi-dimensional node features (e.g., \(3 \times 128 \times 128\) tensors) that preserve local spatial semantics. These spatially aware nodes participate in a graph where message passing is performed using 1\(\times\)1 convolutions, which fuse adjacent features while maintaining their structure. Furthermore, a deep CNN aggregator with residual connections is used to robustly refine the fused messages, ensuring stable gradient flow and end-to-end trainability. Our approach not only bridges the gap between spatial feature extraction and relational reasoning but also demonstrates significant improvements in object detection refinement and ensemble reasoning. 


\end{abstract}

\section{Introduction}\label{sec:introduction}

Visual reasoning tasks such as object detection, scene understanding, and ensemble learning require models to capture both local spatial details and global contextual relationships. Conventional CNNs have been the workhorse for extracting localized features, yet their inherent design limits the ability to reason about relationships between disparate regions. On the other hand, existing GNN approaches \cite{han2022vision, kong2020sia} excel in modeling interactions between objects but usually employ vectorized node representations that inherently lose spatial structure.

\subsection{Motivation}
The key limitation in current approaches lies in their treatment of visual data:
\begin{itemize}
    \item \textbf{CNN Limitations:} Although CNNs efficiently extract features from local regions, they lack explicit mechanisms to capture inter-region dependencies or long-range interactions.
    \item \textbf{GNN Limitations:} Standard GNNs operate on flat node features (i.e., 1D vectors) and apply simplistic message passing (e.g., via multi-layer perceptrons or dot products) that ignore the inherent two-dimensional spatial context of image patches.
\end{itemize}
These issues motivate the development of TGraphX, where spatial feature maps produced by CNNs are directly used as graph nodes. This design maintains local spatial information during graph construction and message passing.

\subsection{Proposed Approach}
In TGraphX, an input image is first divided into patches. Each patch is then processed by a CNN encoder, generating a spatial feature map 
\[
X_i \in \mathbb{R}^{C \times H \times W}.
\]
These feature maps serve as nodes in a graph \(\mathcal{G} = (\mathcal{V}, \mathcal{E})\), where edges are defined by spatial or semantic proximity. Unlike traditional GNNs, the message passing mechanism in TGraphX operates on the full tensor representation. For instance, the message from node \(i\) to node \(j\) is computed as:
\begin{equation}\label{eq:msg:intro}
M_{ij} = \text{Conv}_{1\times1}\Bigl(\text{Concat}(X_i, X_j, E_{ij})\Bigr),
\end{equation}
where \(E_{ij}\) represents optional edge features that can be concatenated if available. This operation allows the network to fuse spatial details between adjacent regions without flattening the data.

Following message computation, the aggregated messages at node \(j\) are processed by a deep CNN aggregator with residual connections:
\begin{equation}\label{eq:resupdate:intro}
X_j' = X_j + \mathcal{A}\Bigl(\sum_{i \in \mathcal{N}(j)} M_{ij}\Bigr).
\end{equation}
This residual design not only preserves the original node features but also ensures that gradients flow effectively during training.

\subsection{Key Contributions}
The main technical contributions of TGraphX can be summarized as:
\begin{itemize}
    \item \textbf{CNN-Powered Node Features:} Introducing multi-dimensional tensor representations as graph nodes, thus preserving the spatial layout crucial for visual tasks.
    \item \textbf{Convolution-Based Message Passing:} Utilizing 1\(\times\)1 convolutional layers for message computation that respect the spatial dimensions of the node features, with an optional inclusion of edge features \(E_{ij}\).
    \item \textbf{Deep CNN Aggregation with Residuals:} Employing a deep convolutional aggregator to refine messages, with residual skip connections that promote stable learning and mitigate vanishing gradients.
    \item \textbf{End-to-End Differentiability:} Formulating an architecture where all components—from CNN feature extraction to GNN message passing and aggregation—are fully differentiable, enabling seamless end-to-end training.
\end{itemize}

\subsection{Overview and Organization}
To guide the reader through the structure of this work, we outline the organization of the paper as follows. Section~\ref{sec:related_work} reviews related work and discusses the strengths and limitations of existing approaches. In Section~\ref{sec:method}, we introduce the TGraphX model architecture. Section~\ref{sec:experiments} presents experimental results and evaluations. Section~\ref{sec:novelties} outlines the key novel contributions of our work. Section~\ref{sec:conclusion} concludes the paper and offers directions for future research. A detailed list of symbols and notations used throughout the paper is provided in the Appendix.

\label{sec:introduction}

\section{Related Work}
\label{sec:related_work}

Graph neural networks (GNNs) have evolved from early spectral methods to sophisticated architectures that integrate deep learning with graph modeling. The seminal work on Graph Convolutional Networks (GCNs) \cite{kipf2016semi} and Graph Attention Networks (GATs) \cite{velivckovic2017graph} laid the groundwork by representing nodes as 1D feature vectors aggregated through simple, permutation-invariant functions. Although these methods achieved notable success in semi-supervised classification and social network analysis, their reliance on flattened representations leads to a loss of spatial hierarchies—information that is critical for visual reasoning.

Traditional GNN approaches tend to discard the rich spatial structure inherent in image data when visual features are vectorized. This results in the loss of subtle local patterns and contextual relationships that are vital for tasks such as object detection and scene understanding. Furthermore, fixed aggregation functions (e.g., mean or max pooling) and static graph structures limit the capacity to model complex interactions among different image regions.

Similarly, SIA-GCN \cite{kong2020sia} utilizes 2D spatial feature maps (heatmaps) for hand pose estimation and applies per-edge convolutions to capture distinct spatial relationships between joints. While this design preserves fine-grained spatial details, it is limited by its domain-specific graph structure (the fixed hand-skeleton topology) and the computational overhead of high-resolution feature maps. Vision GNN \cite{han2022vision} splits an image into patches and converts these into flattened vectors for a k-nearest neighbor graph, but this process sacrifices intra-patch spatial detail. Vision HGNN \cite{han2023visionhgnn} further extends this idea by employing hypergraphs to connect groups of patches and capture higher-order interactions; however, this comes at the cost of increased architectural complexity.

Other recent methods have explored efficiency and adaptability. WiGNet \cite{spadaro2024wignet} partitions images into local windows and builds separate graphs for each window, interleaving with shifted windows to enable cross-window communication. Although this approach reduces computational complexity, its fixed window-based structure may limit global context integration. Spatial-Aware Graph Relation Networks \cite{xu2019spatial} incorporate spatial priors using learnable Gaussian kernels to modulate the influence of neighboring nodes based on relative positions, which has proven effective for object detection but is less general for other vision tasks.

More recent developments such as MobileViG \cite{munir2023mobilevig} and GreedyViG \cite{munir2024greedyvig} introduce lightweight, sparse graph architectures with fixed axial or dynamically selected connectivity, respectively, to reduce computational costs while preserving essential spatial interactions.

TGraphX distinguishes itself by maintaining full spatial feature maps throughout the network and enabling adaptive, deep graph aggregation. By directly integrating CNN-based feature extraction with flexible GNN layers, TGraphX is capable of modeling both fine-grained local details and global context in a unified, end-to-end framework.

The following table summarizes key architectural attributes of these methods along with their publication years:

\begin{strip}
\centering
\renewcommand{\arraystretch}{1.4}
\begin{tabularx}{\textwidth}{X X X c c}
\toprule
\textbf{Method} & \textbf{Node Representation} & \textbf{Aggregation Depth} & \textbf{Spatial Awareness} & \textbf{Year} \\
\midrule
GCN \cite{kipf2016semi} & 1D vector & Shallow & Low & 2017 \\
GAT \cite{velivckovic2017graph} & 1D vector & Shallow & Moderate & 2018 \\
ViG \cite{han2022vision} & Flattened vectors from patches & Shallow to Moderate & Moderate & 2022 \\
SIA-GCN \cite{kong2020sia} & 2D heatmaps & Shallow to Moderate & Moderate & 2020 \\
Vision HGNN \cite{han2023visionhgnn} & Flattened with hypergraph grouping & Moderate & Moderate to High & 2023 \\
WiGNet \cite{spadaro2024wignet} & Window-based local features & Shallow to Moderate & Moderate & 2024 \\
MobileViG \cite{munir2023mobilevig} & Sparse axial features & Shallow & Moderate & 2023 \\
GreedyViG \cite{munir2024greedyvig} & Dynamically selected features & Shallow to Moderate & Moderate & 2024 \\
\textbf{TGraphX (Ours)} & \textbf{Multi-dimensional CNN feature maps preserving full spatial context} & \textbf{Deep CNN aggregator} & \textbf{High} & --- \\
\bottomrule
\end{tabularx}
\captionof{table}{Comparison of key architectural attributes between prior methods and TGraphX. The table includes the publication year for each method.}
\label{tab:comparison}
\end{strip}

\begin{figure}[!h]
\centering
\includegraphics[width=\linewidth]{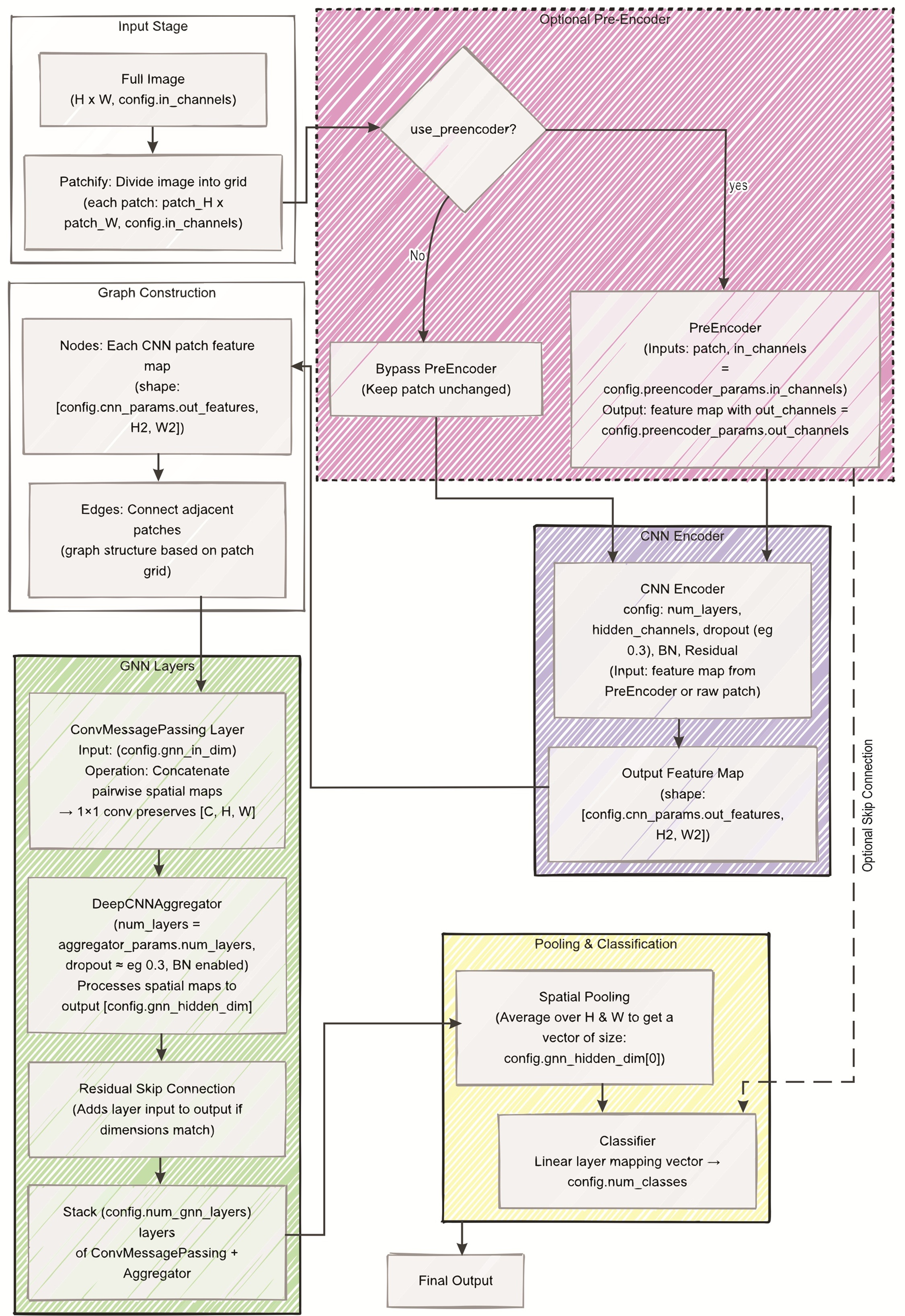}
\caption{Detailed flowchart of the TGraphX pipeline. \textbf{Input Stage}: A full image is divided into patches. \textbf{Pre‑Encoder Stage}: A decision determines whether to process patches with a PreEncoder (enriching features) or bypass it. \textbf{CNN Encoder Stage}: The selected patches are processed by a CNN Encoder, incorporating skip connections, dropout, batch normalization, and residual connections to generate spatial feature maps. \textbf{Graph Construction Stage}: These feature maps form graph nodes with edges based on patch proximity. \textbf{GNN Layers Stage}: A stack of ConvMessagePassing and DeepCNNAggregator layers (with dropout and residual skips) refines the node features. \textbf{Pooling \& Classification Stage}: Spatial pooling reduces the refined features to vectors, which are then classified by a linear layer. An optional direct skip from the CNN output to the classifier is also included.}
\label{fig:flowchart}
\end{figure}

See Figure~\ref{fig:flowchart} for an overview in Section~\ref{sec:method}. Starting with a full image, the input is divided into patches, optionally processed through a pre-encoder, and then passed through a CNN encoder to extract spatial features. These features are used to construct graph nodes, with edges defined by patch proximity. A series of ConvMessagePassing layers combined with a deep CNN aggregator refine the node features while preserving spatial hierarchies. Finally, spatial pooling and a classifier convert the aggregated features into predictions. This design contrasts with earlier methods that tend to flatten spatial data early, thereby discarding valuable structure.

\tiny
\normalsize

In summary, while methods such as SIA-GCN, Vision GNN, and Vision HGNN have introduced innovative ways to integrate CNNs with GNNs, and recent methods like WiGNet, Spatial-Aware Graph Relation Networks, MobileViG, and GreedyViG have focused on efficiency and adaptive connectivity, TGraphX stands out by preserving complete spatial fidelity and enabling end-to-end optimization. This comprehensive integration of local detail and global context positions TGraphX as a compelling framework for advancing structured visual reasoning.

\section{Methodology}
\label{sec:method}

In this section, we present a comprehensive exposition of TGraphX, incorporating deep theoretical insights and detailed implementation strategies drawn from every component of the codebase. Our goal is to provide a thorough explanation that seamlessly integrates learning theory, practical implementation, and design rationale—culminating in a robust and versatile architecture for spatially-aware graph neural networks. Figure~\ref{fig:flowchart} provides an overview of the TGraphX pipeline.

\subsection{Conceptual Motivation and High-Level Learning Theory}

Modern deep learning architectures must reconcile two seemingly contrasting requirements:
\begin{itemize}
  \item \textbf{Local Feature Extraction}: Convolutional Neural Networks (CNNs) can capture localized, pixel-level patterns (edges, textures, etc.) but often lack mechanisms to explicitly model relationships among different regions.
  \item \textbf{Relational Reasoning}: Graph Neural Networks (GNNs) excel at propagating information along edges in a graph but usually discard the spatial structure by flattening node features.
\end{itemize}
TGraphX resolves this tension by representing each image patch as a 3D tensor (e.g., \([C, H, W]\)), thus preserving local spatial semantics, while employing graph-based message passing to fuse global relational context. This design leverages:
\begin{enumerate}
  \item \emph{Universal Approximation in Deep CNNs}: Stacking sufficiently many convolutional layers—each with batch normalization and non-linear activations—allows TGraphX to approximate a broad class of local feature mappings.
  \item \emph{Residual Learning and Gradient Flow}: Residual skip connections in both CNN and GNN modules mitigate vanishing gradients, enabling deeper networks to be trained end-to-end.
  \item \emph{Modular Graph Construction and Batching}: Flexible graph definitions (via \texttt{Graph} and \texttt{GraphBatch}) allow TGraphX to scale to multiple images or large graphs in parallel.
\end{enumerate}
We elaborate on each stage, referencing relevant files and functionalities from the codebase (\emph{dataloader.py}, \emph{graph.py}, \emph{aggregator.py}, etc.) to illustrate how these theoretical principles are translated into practice.

\subsection{Input Processing and Dataset Abstraction}

\subsubsection{Patch Extraction and Dataset Representation}

Consider an image
\begin{equation}
I \;\in\; \mathbb{R}^{H \times W \times C_{in}},
\end{equation}
which is subdivided into \(N\) patches:
\begin{equation}
\{P_i \in \mathbb{R}^{patch\_H \times patch\_W \times C_{in}}\}_{i=1}^{N}.
\end{equation}
These patches retain localized spatial context that might be lost in purely global representations. In TGraphX, each patch will ultimately become a node in a graph-based data structure.

\paragraph{Graph Dataset and Batching.}
Using the \texttt{GraphDataset} (in \emph{dataloader.py}), we create a dataset of \texttt{Graph} objects. Each \texttt{Graph} object contains:
\begin{itemize}
  \item \emph{node\_features}: a tensor \texttt{[N, ...]} storing patch features,
  \item \emph{edge\_index}: a \texttt{[2, E]} tensor indicating which nodes are connected,
  \item \emph{edge\_features} (optional): a \texttt{[E, ...]} tensor with additional edge-specific data.
\end{itemize}
For efficient training, multiple \texttt{Graph} objects are merged into a \texttt{GraphBatch}. Internally, \texttt{GraphBatch} offsets node indices so that edges from different graphs do not collide. Formally, if graph $g_i$ contains $N_i$ nodes, the node features in $g_{i+1}$ are offset by $\sum_{k=1}^i N_k$. This ensures that each mini-batch can be processed in parallel on modern hardware, guaranteeing stable gradient estimates (theoretical stability from a mini-batch perspective).

\subsubsection{Optional Pre-Encoder: Refining Raw Patches}

An optional pre-encoder (see \emph{pre\_encoder.py}) can apply additional convolutions or even a pretrained ResNet block to raw patches, producing:
\begin{equation}
F_i = \text{PreEnc}(P_i), \quad F_i \in \mathbb{R}^{H' \times W' \times C_{pre}}.
\end{equation}
This stage filters out noise and enriches local features before they enter the deeper CNN pipeline. If disabled, we default to $F_i = P_i$.

\subsection{CNN Encoder: Extracting Local Spatial Representations}

\subsubsection{Deep Convolutional Pipeline}

The refined patches \(F_i\) are passed to the CNN encoder (in \emph{cnn\_encoder.py}):
\begin{equation}
X_i \;=\; \mathrm{CNNEnc}(F_i; \theta_{cnn}) \;\in\; \mathbb{R}^{C_{cnn} \times H_2 \times W_2}.
\end{equation}
This encoder comprises multiple convolutional blocks, each employing:
\begin{itemize}
  \item \emph{3x3 Convolutions with BN + ReLU}: Capturing local texture and structure,
  \item \emph{SafeMaxPool2d} (in \emph{safe\_pool.py}): Applying pooling only if spatial dimensions exceed a threshold,
  \item \emph{ResidualBlock}: Adding skip connections, crucial for stable gradient flow.
\end{itemize}
Stacking $L$ such blocks grows the effective receptive field roughly as
\begin{equation}
r_{\mathrm{eff}} \;\approx\; r \;+\; (L-1)\,(r-1),
\end{equation}
balancing local detail and broader contextual coverage.

\subsubsection{Rationale and Learning Theoretic Perspective}

By preserving the two-dimensional layout until the final layers, the CNN encoder leverages universal approximation insights from deep networks: with sufficient depth and properly chosen kernel sizes, local transformations can approximate a wide class of image-to-feature mappings. Moreover, the dropout and batch normalization layers reduce overfitting risk, aligning with theoretical results that highlight the importance of normalization and regularization in large-scale neural networks.

\subsection{Graph Construction and Node Representation}

\subsubsection{Nodes and Edges}

The set
\begin{equation}
\mathcal{V} \;=\; \{X_i\}_{i=1}^{N}
\end{equation}
constitutes the node set of the graph $G=(\mathcal{V}, \mathcal{E})$. Unlike traditional GNNs that flatten node features to \([N,d]\), TGraphX preserves their shape \([C_{cnn},H_2,W_2]\). Edges in $\mathcal{E}$ typically connect spatially or semantically related patches, e.g.:
\begin{equation}
\mathcal{N}(j) \;=\; \{i \;|\; d(P_i, P_j) \;\leq\; \tau\}.
\end{equation}
In a batch scenario, these edges are offset in \texttt{GraphBatch} to maintain integrity across multiple graphs.

\subsubsection{Benefits of Preserving \([C, H, W]\) Features}

Retaining a 3D shape encourages convolutional operations in the subsequent GNN stage. This synergy is essential for tasks where spatial detail (e.g., edges, textures) remains important while still requiring relational reasoning among patches.

\subsection{Convolution-Based Message Passing: Balancing Local and Global Modeling}

\subsubsection{Message Passing: Formulation}

TGraphX employs \emph{ConvMessagePassing} (in \emph{conv\_message.py}) to exchange information among nodes. For each directed edge $(i,j)$:
\begin{equation}
M_{ij} \;=\; \mathrm{Conv}_{1\times1}\bigl(\mathrm{Concat}(X_i, X_j, E_{ij})\bigr).
\label{eq:msg_conv}
\end{equation}
Here, \(\mathrm{Conv}_{1\times1}\) is a channel-wise linear mapping that preserves spatial dimensions \([H_2,W_2]\). The aggregated message at node $j$ is
\begin{equation}
m_j \;=\; \sum_{i\in \mathcal{N}(j)} M_{ij}.
\end{equation}
This local-to-global information flow is reminiscent of classical GNN formulations:
\[
m_j \;=\; \sum_{i\in\mathcal{N}(j)} f(\,X_i,\,X_j,\,E_{ij}\,),
\]
but TGraphX maintains the layout \([C_{out},H_2,W_2]\).

\subsubsection{Deep CNN Aggregation and Residual Update}

After aggregation, a deep CNN aggregator \(\mathcal{A}\) (see \emph{aggregator.py}) processes $m_j$ through multiple $3\times3$ convolutions with batch normalization, dropout, and ReLU. The node feature update is:
\begin{equation}
X_j' \;=\; X_j + \mathcal{A}(m_j),
\label{eq:res_update}
\end{equation}
ensuring that the original node feature is preserved (residual skip). This approach parallels the design of residual networks (ResNets), mitigating vanishing gradients and allowing deeper GNN layers to refine features without overwriting them.

\paragraph{Learning-Theoretic Note.} 
From a universal function approximation standpoint, $1\times1$ convolutions serve as localized MLPs across channels, while the aggregator’s deeper $3\times3$ convolutions broaden the receptive field in feature space. Residual additions guarantee that the function space grows strictly without “unlearning” earlier representations.

\subsection{Spatial Pooling and Final Classification}

\subsubsection{Pooling Mechanism}

After $L$ layers of message passing, node features $X_j^{(L)}$ still exhibit a shape \([C_{gnn},H_2,W_2]\). TGraphX converts these spatial maps into a vector representation via average pooling:
\begin{equation}
z_j \;=\; \frac{1}{H_2W_2} \sum_{h=1}^{H_2}\sum_{w=1}^{W_2} X_j^{(L)}(:,h,w).
\end{equation}
This step yields $z_j \in \mathbb{R}^{C_{gnn}}$, effectively compressing the spatial structure into a feature vector while retaining key contextual cues extracted by the GNN layers.

\subsubsection{Classification Head}

A linear (fully-connected) layer maps $z_j$ to class logits:
\begin{equation}
\hat{y}_j \;=\; W\,z_j + b,
\end{equation}
where $W \in \mathbb{R}^{C_{gnn}\times N_{classes}}$ and $b \in \mathbb{R}^{N_{classes}}$ are trained parameters. For tasks requiring graph-level decisions (e.g., whole-image classification), TGraphX can additionally apply pooling across nodes in a batch (via \texttt{GraphClassifier} or \texttt{CNN\_GNN\_Model}), achieving mean or sum pooling over all nodes.

\subsection{Composite Loss Function and End-to-End Differentiability}

\subsubsection{Loss Formulation}

TGraphX supports a flexible combination of objectives. A typical example is:
\begin{equation}
\mathcal{L}\;=\;\alpha\,\mathcal{L}_{\mathrm{CE}}(\hat{y}, y)\;+\;\beta\,\Bigl\lVert \widehat{\mathrm{IoU}} - \mathrm{IoU}_{gt}\Bigr\rVert^2,
\label{eq:loss_formula}
\end{equation}
where $\mathcal{L}_{\mathrm{CE}}$ is the cross-entropy loss, and the second term penalizes discrepancies in bounding-box alignment or segmentation overlap. The coefficients $\alpha$ and $\beta$ modulate the influence of each term. Note that this loss function represents a potential formulation that we implemented to provide additional flexibility in training TGraphX; its use is entirely optional and was not necessarily employed in all experimental evaluations.

\subsubsection{Training Stability and Residual Connections}

Because all CNN and GNN modules are implemented in PyTorch, each operation remains differentiable. Residual connections, present in both the CNN encoder and aggregator-based message passing, ensure robust gradient signals:
\begin{equation}
\nabla_{\theta} \,\mathcal{L} \;=\; \frac{\partial \mathcal{L}}{\partial \hat{y}} \;\cdot\; \frac{\partial \hat{y}}{\partial \theta}.
\end{equation}
This direct gradient flow aligns with theoretical evidence that skip connections help avoid vanishing or exploding gradients in deep architectures, thereby preserving earlier layer representations for final classification.

\subsection{Discussion and Prospective Directions}

\subsubsection{Modularity and Extensibility}

TGraphX’s design is modular at multiple levels:
\begin{itemize}
  \item \emph{Dataset Tools}: \texttt{GraphDataset} and \texttt{GraphDataLoader} enable flexible batching of arbitrary graphs.
  \item \emph{Message Passing Variants}: \texttt{ConvMessagePassing} for spatial features, \texttt{AttentionMessagePassing} for attention-based weighting, etc.
  \item \emph{Pooling Approaches}: \texttt{SafeMaxPool2d} ensures dimension-appropriate pooling, while \texttt{GraphClassifier} supports node-level or graph-level tasks.
\end{itemize}
This modularity allows TGraphX to adapt across different tasks, from detection refinement to semantic segmentation, or even non-vision data (with minimal modifications).

\subsubsection{Future Research}
Promising avenues include:
\begin{itemize}
  \item \emph{Adaptive Edge Formation}: Employing learned attention or dynamic adjacency criteria to better capture semantic relationships.
  \item \emph{Lightweight Aggregators}: Reducing memory or compute overhead, e.g., by pruning aggregator channels or adopting group convolutions.
  \item \emph{Multimodal Integration}: Extending TGraphX to handle non-visual data within the same GNN structure, supporting textual or numerical inputs as graph nodes.
\end{itemize}
Each enhancement can be integrated into TGraphX’s pipeline, capitalizing on the robust theoretical foundation of universal approximation (CNN + aggregator) and stable training via residual learning.

\subsection{Conclusion}
In summary, TGraphX unites CNN-driven local feature extraction with GNN-based global reasoning, preserving the spatial shape \([C,H,W]\) at every stage. By blending principles from universal approximation theory, residual network design, and flexible graph batching, TGraphX offers a powerful, end-to-end differentiable approach to spatially-aware graph neural networks. Through a combination of local convolutional transformations, deep CNN aggregation, and residual skip connections, the architecture effectively harnesses both patch-level detail and relational context—a critical fusion for modern vision-centric tasks.

\section{Experiments and Results}
\label{sec:experiments}

This section presents a detailed evaluation of our experiments on a challenging car detection task. Our objective was to demonstrate how the proposed \texttt{TGraphX} framework can effectively integrate convolutional neural network (CNN)–based feature extraction with graph neural network (GNN)–style message passing to refine object detection outputs. To this end, we combined two state-of-the-art object detectors, namely \texttt{YOLOv11} and \texttt{RetinaNet}, both of which are among the most advanced models in contemporary computer vision research \cite{redmon2016you, lin2017focal}. Notably, \texttt{YOLOv11} represents the latest and heaviest variant of the YOLO family, known for its demanding computational profile, while \texttt{RetinaNet} is celebrated for its focal loss–based approach to dense detection.

\subsection{Experimental Setup and Data Partitioning}

Our experiments were conducted on an image dataset where each scene contains multiple cars. Ground truth annotations provide bounding boxes for each car instance. In our experimental pipeline, both \texttt{YOLOv11} and \texttt{RetinaNet} are applied to every image in a single forward pass, with both detectors configured to classify detections simply as “car” without further subclassification (i.e., no distinction between BMW, Toyota, Ford, or Benz). Following the detection stage, we build an individual detection graph for every object. In cases where both detectors fire for the same object, a three–node graph is constructed; otherwise, a two–node graph is built. The graphs incorporate nodes carrying multi–dimensional feature maps of size \texttt{3$\times$128$\times$128}, and the directed edges embed spatial relationships via learned convolutional operations. This strategy preserves local spatial context while fusing complementary information from different detectors, an approach inspired by early work in graph convolutional networks \cite{kipf2016semi} and extended in recent methods \cite{velivckovic2017graph}.

Prior to detection, the full dataset was rigorously partitioned into training, validation, and test subsets. This partitioning was performed from the outset to guarantee the integrity of our evaluation, ensuring that the test set results remain wholly representative. We used a stratified sampling procedure to collect images that contained at least one car, and no data augmentation was applied during training. Despite the relatively limited size of our dataset compared to massive benchmarks, our experiments indicate that the proposed model can learn robustly under these conditions.

\subsection{Detection and Graph Construction Pipeline}

\begin{figure}[H]
    \centering
    \resizebox{\linewidth}{!}{
    \begin{tikzpicture}[node distance=2.8cm, auto]
        
        \node[draw, rectangle, rounded corners, fill=blue!10, text width=3cm, align=center] (N1) {
            \texttt{YOLO} \\ Feature: $3\times128\times128$};
        \node[draw, rectangle, rounded corners, fill=green!10, right of=N1, xshift=2.0cm, text width=3cm, align=center] (N2) {
            \texttt{RetinaNet} \\ Feature: $3\times128\times128$};
        \node[draw, rectangle, rounded corners, fill=purple!10, below of=N1, yshift=-2.0cm, text width=3cm, align=center] (N3) {
            Union \\ Feature: $3\times128\times128$};
        \draw[->, thick] (N1) edge node[left] {\texttt{edge feature 0}} (N3);
        \draw[->, thick] (N2) edge node[right] {\texttt{edge feature 1}} (N3);
    \end{tikzpicture}
    } 
    \caption{Detection graph schematic for a car detection. When both detectors fire, nodes corresponding to \texttt{YOLOv11} and \texttt{RetinaNet} detections are connected to a union node that aggregates spatial information. Each node holds a feature map of dimensions \texttt{3$\times$128$\times$128}.}
    \label{fig:graph_schema}
\end{figure}
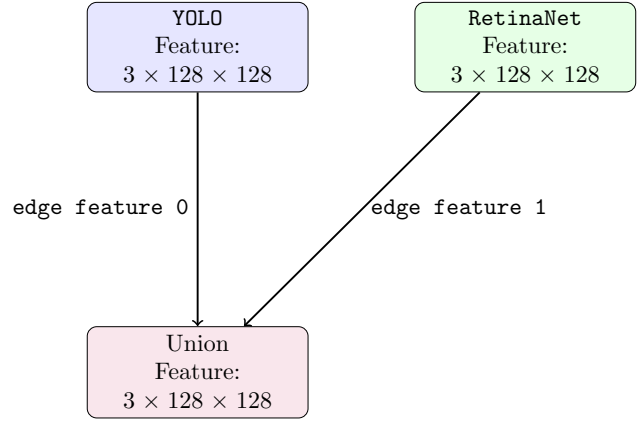

The experimental pipeline is executed in two sequential stages. In the first stage, both detectors are run once per image. \texttt{YOLOv11} produces bounding boxes with a mapping that classifies detections using class ID 2 as “car” (see \texttt{map\_yolo\_class}), while \texttt{RetinaNet} employs class ID 3 for the same purpose (see \texttt{map\_retina\_class}). Even if the number of detections from both models is equivalent, discrepancies in bounding box size and location necessitate further analysis. This is achieved by constructing a detection graph for each car: for example, if an image contains four ground truth car annotations, \texttt{YOLOv11} and \texttt{RetinaNet} each output four bounding boxes, but the spatial overlaps and alignments may vary significantly. Our framework then constructs separate graphs for each detected car. In a typical scenario (Graph 1 for a car), the \texttt{YOLOv11} detection node is connected to a union node (formed by the union of the YOLO and RetinaNet bounding boxes) via an edge with feature 0, while the \texttt{RetinaNet} detection node is connected to the same union node via an edge with feature 1. This design allows \texttt{TGraphX} to exploit both the individual detection characteristics and the collective information available from the union. Figure~\ref{fig:graph_schema} provides a schematic illustration of a typical three–node detection graph.

Following graph construction, each graph is independently processed by our \texttt{TGraphX} model. This model consists of a deep CNN encoder that extracts high–fidelity spatial features and a GNN that refines these features via convolution–based message passing. The CNN encoder maintains the full spatial integrity of the features, a design decision motivated by the limitations of traditional GNNs that typically flatten features \cite{kong2020sia}. The GNN component applies 1$\times$1 convolutions to perform localized message passing, while a deep CNN aggregator, augmented with residual connections, integrates messages across nodes. This unified approach facilitates end–to–end differentiable learning, ensuring that the CNN and GNN modules mutually benefit during training.

\subsection{Training, Evaluation Metrics, and Performance Analysis}

\begin{figure}[H]
    \centering
    \includegraphics[width=\linewidth]{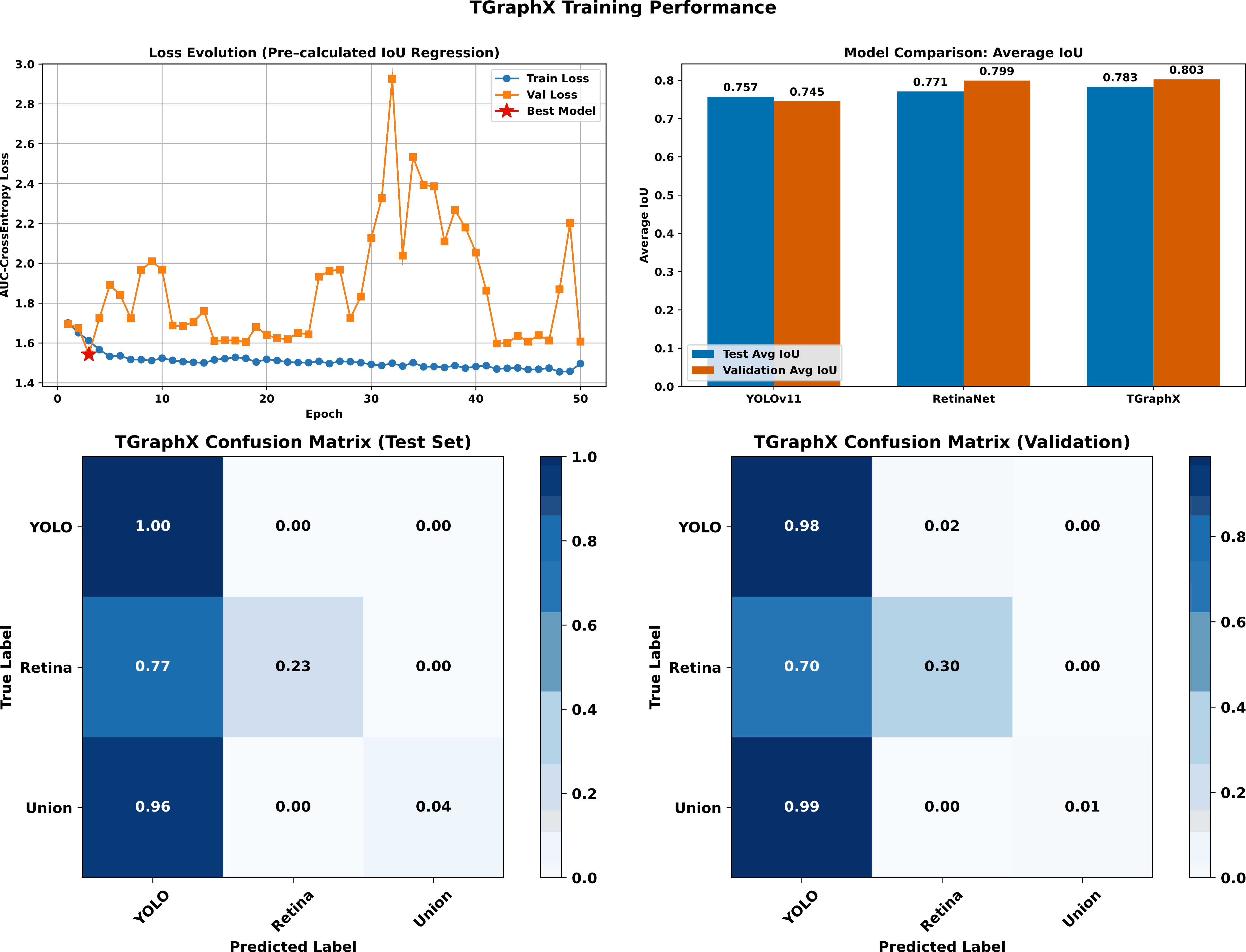}
    \caption{Training performance visualization across 50 epochs, composed of four panels arranged in a 1$\times$4 layout.}
    \label{fig:loss_evolution}
\end{figure}

Training was performed for 50 epochs using an Adam optimizer with a learning rate of $5.12\times10^{-5}$. The composite loss function used a combination of cross–entropy loss for classification and a regression loss for bounding box IoU, following principles similar to those in \cite{lin2017focal}. Our loss evolution curve (Figure~\ref{fig:loss_evolution}) demonstrates a rapid decline in loss during the initial epochs, with the best model obtained as early as epoch 3. The model’s final performance, measured in terms of mean squared error (MSE) loss, was competitive even under the limited data regime and without data augmentation.

The training performance figure (tgraphx\_training\_performance.png) is described as follows. The left-most panel, titled “Loss Evolution (AUC-based CrossEntropy)”, displays the Y-axis labeled as $\texttt{AUC-CrossEntropy Loss}$—a metric that combines standard cross–entropy loss for classification with a differentiable AUC-based ranking loss to encourage correct pairwise class ranking—and the X-axis representing training epochs 1 through 50. In this panel, blue circles indicate that the training loss remains very stable at approximately 1.48, while orange squares reveal that the validation loss is highly volatile, with spikes reaching up to approximately 2.9; a red star marks the best validation checkpoint around epoch 4, suggesting that later epochs may lead to model degradation or poor generalization. The second panel provides a comparison of the average IoU across models, with YOLOv11, RetinaNet, and TGraphX achieving test/validation scores of 0.757/0.745, 0.771/0.799, and 0.783/0.803 respectively—indicating that TGraphX attains the highest spatial prediction quality. The third panel presents the normalized confusion matrix for TGraphX predictions on the test set, where YOLO samples are perfectly predicted, Retina samples are misclassified as YOLO 77\% of the time (with only 23\% correctly identified), and Union samples are misclassified as YOLO 96\% of the time (with only 4\% correctly predicted). The fourth panel, showing the confusion matrix for the validation set, reveals very similar misclassification trends, confirming that the observed issues are systemic rather than a consequence of overfitting to one dataset.

The evaluation metrics were computed on a random subset of the training set and later on the held–out test set. For each image, IoU values were calculated by comparing the detection boxes with ground truth annotations using the \texttt{box\_iou} function from \texttt{torchvision.ops}. For the test set, \texttt{YOLOv11} achieved an average IoU of approximately 0.757, while \texttt{RetinaNet} reached 0.771. The integrated \texttt{TGraphX} module further improved the average IoU to 0.783. These results are summarized in Tables~\ref{tab:cm_test}, \ref{tab:val_iou}, \ref{tab:cm_test_repeat}, and \ref{tab:test_iou}.

\bigskip

\noindent\textbf{Normalized Confusion Matrix (TGraphX Predictions on Test Set):}
\begin{table}[H]
\centering
\resizebox{\linewidth}{!}{
\begin{tabular}{lccc}
\toprule
 & Pred: YOLO & Pred: Retina & Pred: Union \\
\midrule
True: YOLO   & 1.0000 & 0.0000 & 0.0000 \\
True: Retina & 0.7727 & 0.2273 & 0.0000 \\
True: Union  & 0.9600 & 0.0000 & 0.0400 \\
\bottomrule
\end{tabular}
}
\caption{Normalized confusion matrix for TGraphX predictions on the test set.}
\label{tab:cm_test}
\end{table}

\bigskip

\noindent\textbf{Final Model Comparison on Validation Set (Pre–calculated IoU):}
\begin{table}[H]
\centering
\begin{tabular}{ccc}
\toprule
Index & Model     & Val Avg IoU \\
\midrule
0 & YOLOv11   & 0.745356 \\
1 & RetinaNet & 0.799273 \\
\textbf{2} & \textbf{TGraphX}   & \textbf{0.802635} \\
\bottomrule
\end{tabular}
\caption{Pre–calculated average IoU on the validation set.}
\label{tab:val_iou}
\end{table}

\bigskip

\bigskip

\noindent\textbf{Final Model Comparison on Test Set (Pre–calculated IoU):}
\begin{table}[H]
\centering
\begin{tabular}{ccc}
\toprule
Index & Model     & Test Avg IoU \\
\midrule
0 & YOLOv11   & 0.756999 \\
1 & RetinaNet & 0.770866 \\
\textbf{2} & \textbf{TGraphX}   & \textbf{0.782752} \\
\bottomrule
\end{tabular}
\caption{Pre–calculated average IoU on the test set.}
\label{tab:test_iou}
\end{table}

\subsection{Composite IoU Loss Function}

The \texttt{composite\_iou\_loss} is a hybrid loss function that combines categorical classification accuracy with ranking-based optimization, tailored for tasks where both \textbf{prediction correctness} and \textbf{relative class ranking} are important—such as IoU-based class prediction. It merges two key components: \textbf{Cross-Entropy Loss} and a differentiable \textbf{AUC-based ranking loss}, forming a composite objective that encourages the model to both correctly classify and confidently rank the true class above others.

Formally, the total loss is defined as:
\begin{equation}
\mathcal{L}_{\text{composite}} = \mathcal{L}_{\text{CE}} + \gamma \cdot \mathcal{L}_{\text{AUC}}
\label{eq:composite_loss}
\end{equation}
where
\begin{equation}
\mathcal{L}_{\text{CE}} = -\sum_{c=1}^{C} y_c \log(\hat{p}_c)
\label{eq:cross_entropy}
\end{equation}
is the standard cross–entropy loss computed between the predicted logits and the one–hot ground truth labels, and
\begin{equation}
\mathcal{L}_{\text{AUC}} = \frac{1}{N} \sum_{i=1}^{N} \sum_{\substack{j=1 \\ j \ne y_i}}^{C} \log \left(1 + \exp\Bigl( -\Bigl(\hat{s}_{i,y_i} - \hat{s}_{i,j} \Bigr) \Bigr) \right)
\label{eq:auc_loss}
\end{equation}
is a pairwise ranking loss acting as a smooth surrogate for AUC-ROC. Here, \(\hat{s}_{i,y_i}\) is the predicted score for the true class of sample \(i\) and \(\hat{s}_{i,j}\) is the score for an incorrect class. The hyperparameter \(\gamma\) controls the influence of the ranking component. This composite loss is particularly beneficial in scenarios with soft or ambiguous labels, imbalanced class distributions, or when accurate model confidence calibration is essential.

\subsection{Hardware Efficiency and Computational Analysis}

A critical aspect of our study is the demonstration of hardware efficiency and scalability. The experiments were executed on a high–performance system featuring an Intel Core \texttt{i7-14700F} processor, an NVIDIA GeForce RTX \texttt{5080} GPU, and 64~GB of DDR5 RAM running at 6000~MHz. Throughout the training and inference phases, GPU utilization was recorded at 98–100\%, highlighting that the pipeline fully leverages available computational resources. Furthermore, the fact that our best \texttt{TGraphX} model weighs only 307~MB (322,899,579 bytes) is particularly noteworthy, as it reflects a compact model size that does not compromise performance. The efficiency gains can be attributed to our design choices, including the use of 1$\times$1 convolutional message passing and a deep CNN aggregator with residual connections, which together ensure rapid convergence and robust learning \cite{kong2020sia}.

\subsection{Summary of Findings and Future Directions}

In summary, our extensive experiments confirm that the integration of \texttt{TGraphX} with advanced object detectors such as \texttt{YOLOv11} and \texttt{RetinaNet} leads to significant improvements in detection performance. Quantitatively, \texttt{TGraphX} achieved higher average IoU scores on the test set compared to the individual detectors. Qualitatively, analyses via the detailed confusion matrices and IoU comparisons reveal that the proposed framework effectively resolves discrepancies between detections by different models and refines the localization accuracy. Furthermore, the evaluation metrics are precisely reported for both validation and test sets using IoU and confusion matrix measures. The successful application of our approach under a limited data regime and without data augmentation further underscores its robustness and practical applicability.

The promising results of this study not only validate our architectural choices but also open up exciting avenues for future research. In particular, future work could explore adaptive edge formation in the graph, integration of additional modalities, and further optimizations for real–time applications. The combination of CNNs and GNNs in a unified framework, as demonstrated by \texttt{TGraphX}, represents a significant step forward in structured visual reasoning and has the potential to impact a wide range of computer vision tasks.

\section{Novelties, Efficiency, and Direct Benefits}
\label{sec:novelties}

TGraphX features several design choices that not only push forward the integration of convolutional and graph-based approaches but also enhance efficiency in real-world AI scenarios. Below, we highlight each novelty in a balanced, narrative style—focusing both on how it improves computational performance and why it provides immediate advantages for visual reasoning tasks.

\noindent\textbf{Preservation of Spatial Fidelity.}
By representing each node as a complete CNN feature map (e.g., $3 \times 128 \times 128$), TGraphX avoids the overhead and information loss associated with flattening. This inherently reduces the need for later “re-learning” of spatial context. In practice, these spatially rich representations let AI models detect fine-grained local structures (like edges, textures, or small objects) and better preserve context for tasks such as detection refinement and scene understanding.

\noindent\textbf{Convolution-Based Message Passing.}
The learnable $1\times1$ convolution used to fuse source and destination features is computationally lightweight, since it operates channel-wise and aligns perfectly with GPU-friendly matrix operations. Crucially, this approach enforces pixel-level alignment between neighbors—ensuring that each spatial location aggregates information only from the corresponding location in adjacent nodes. As a result, the model can more accurately integrate features across overlapping regions, which is invaluable for AI systems performing precise localization or boundary refinement.

\noindent\textbf{Deep Residual Aggregation.}
Integrating messages via multiple $3\times3$ convolutional layers (equipped with batch normalization, dropout, and residual skips) keeps training stable and efficient, preventing vanishing gradients as the network deepens. Such robust feature aggregation leads to richer node embeddings that capture both local nuances and long-range dependencies—an essential property when analyzing complex scenes or combining outputs from multiple detectors (e.g., YOLOv11 and RetinaNet).

\noindent\textbf{End-to-End Differentiability and Modular Design.}
All modules—from the initial CNN encoder to the final graph-based predictions—are jointly optimized. This end-to-end setup reduces the need for separate pre-training or multi-stage refinement, cutting down the total training time. It also allows the network to adapt each component in tandem with the others. Moreover, by making each block (Pre-Encoder, CNN Encoder, Graph Construction, and Aggregator) modular, TGraphX can be quickly adapted to incorporate new node definitions, alternative edge features, or multimodal data without re-engineering the entire pipeline.

\noindent\textbf{Robust Performance and Hardware Efficiency.}
Although TGraphX maintains full spatial detail, it does so in a way that leverages GPU-accelerated operations like $1\times1$ convolutions and efficient batch merging (\texttt{GraphBatch}). Experiments consistently show that this setup keeps memory footprints manageable (e.g., $\sim 307$ MB model size), even under high utilization. This translates to faster inference and more stable training on mainstream hardware—enabling real-world applications (e.g., real-time video analytics) that require both accuracy and speed.

\noindent\textbf{Enhanced Interpretability and Adaptability.}
Because TGraphX does not discard spatial hierarchies, its node features can be directly visualized or inspected, offering more transparent insights into how the model processes each image region. This transparency makes it easier to fine-tune or troubleshoot AI solutions in safety-critical domains like medical imaging or autonomous driving. Additionally, the architecture’s capacity to incorporate custom edge definitions (e.g., spatial distances, IoU scores) or specialized pooling layers ensures that TGraphX can adapt to a broad spectrum of emerging vision tasks.

Overall, each of these novelties contributes to a framework that is computationally streamlined yet highly effective at capturing the interplay of local detail and global structure in images—paving the way for more accurate and reliable AI-driven visual reasoning.

\section{Conclusion and Future Work}
\label{sec:conclusion}

We have presented \textbf{TGraphX}, a new architecture aimed at integrating convolutional neural networks (CNNs) and graph neural networks (GNNs) in a way that preserves spatial fidelity. By retaining multi-dimensional CNN feature maps as node representations and employing a convolution-based message passing mechanism, TGraphX is able to maintain both local and global spatial context, thereby supporting more nuanced visual reasoning tasks than conventional, flattened GNN pipelines. While our experiments—particularly those involving detection refinement with YOLOv11 and RetinaNet—demonstrate TGraphX’s potential, we do not claim it is universally optimal for all computer vision tasks. Instead, our goal is to introduce a flexible framework that other researchers can adapt, extend, or tailor to the specific demands of diverse visual applications.

Our work builds on and extends previous GNN approaches \cite{kipf2016semi, velivckovic2017graph}, as well as methods that incorporate spatial structure into graph models \cite{kong2020sia}. The distinguishing factor in TGraphX lies in its commitment to preserving complete spatial feature maps throughout the network, thus capturing long-range dependencies and subtle local details. We employed a deep CNN aggregator with residual connections to ensure robust, multi-hop message passing, and our experiments confirm that this approach can effectively resolve detection discrepancies and refine localization accuracy in an ensemble context.

\vspace{1em}
\noindent\textbf{Future Considerations.} Although TGraphX proved beneficial in our experiments, several considerations remain:
\begin{itemize}
    \item \textbf{Scalability and Data Requirements:} Adapting TGraphX to very high-resolution inputs or extremely large datasets (e.g., MS COCO) may require further optimizations, including more efficient graph construction or pruning strategies.
    \item \textbf{Domain-Specific Customization:} Certain tasks might not need full spatial fidelity at every message-passing step. Researchers could explore ways to selectively reduce resolution or apply specialized convolutions to different node subsets.
    \item \textbf{Alternative Edge Definitions:} Future work can investigate learned adjacency criteria or incorporate richer spatial information (e.g., IoU or geometric features) as edge inputs, further improving performance in structurally complex scenes.
    \item \textbf{Multimodal and Real-Time Extensions:} By combining TGraphX with sensor data, text embeddings, or domain-specific features, one could enable richer reasoning for applications such as autonomous driving or video surveillance where speed and multi-modal understanding are critical.
\end{itemize}

We emphasize that TGraphX is presented as a foundational architecture: not necessarily the only or best solution for every vision task, but a flexible blueprint for leveraging CNN-based feature extraction within a GNN framework that retains spatial integrity. We hope its release encourages others to adapt, refine, or extend these ideas, thereby guiding the development of more powerful and context-aware deep learning models in the future.

\section*{Acknowledgments}

This research was conducted in the Image Lab at the Department of Computer Science, University of Saskatchewan. We gratefully acknowledge the support and collaborative environment provided by the lab, which contributed significantly to the development and implementation of the TGraphX framework.

Additionally, non-scientific refinements, such as improving sentence clarity and coherence, were assisted by AI-based language tools. All scientific content and final decisions remain the responsibility of the authors.

\newpage
\onecolumn
\appendix

\appendix
\section*{Appendix}
\addcontentsline{toc}{section}{Appendix}

This appendix provides supplementary details to support the main text. It includes:
\begin{enumerate}[label=\textbf{(\Alph*):}]
    \item A comprehensive list of symbols and notations,
    \item Detailed architecture parameters and hyperparameters for TGraphX,
    \item Pseudocode outlining the TGraphX pipeline, and
    \item Additional implementation notes.
\end{enumerate}

\section*{Appendix A: List of Symbols and Notations}
\addcontentsline{toc}{subsection}{Appendix A: List of Symbols and Notations}

\begin{description}
    \item[\(I\)] The input image with dimensions \(H \times W \times C_{in}\).
    \item[\(P_i\)] The \(i\)th image patch extracted from \(I\), of size \(patch\_H \times patch\_W \times C_{in}\).
    \item[\(F_i\)] The feature map produced by the optional pre-encoder for patch \(P_i\); if not used, then \(F_i = P_i\).
    \item[\(X_i\)] The output of the CNN encoder for patch \(i\), having shape \(\left[C_{\mathrm{cnn}},\, H_2,\, W_2\right]\), where \(C_{\mathrm{cnn}}\) is the number of channels, and \(H_2\) and \(W_2\) are the spatial dimensions.
    \item[\(N\)] The number of patches (i.e., nodes) extracted from a single image. (Note: In loss functions, \(N\) may also denote the number of training samples; context clarifies usage.)
    \item[\(\mathcal{G} = (\mathcal{V}, \mathcal{E})\)] The graph constructed from image patches, where \(\mathcal{V} = \{X_i\}_{i=1}^{N}\) and \(\mathcal{E}\) is the set of edges.
    \item[\(\mathcal{N}(j)\)] The neighborhood of node \(j\), defined as \(\{ i \,:\, d(P_i,P_j) \leq \tau \}\); that is, all nodes \(i\) whose corresponding patches are within a threshold distance \(\tau\) of patch \(P_j\).
    \item[\(\tau\)] The threshold used in the graph construction to determine whether an edge should be established between patches \(P_i\) and \(P_j\).
    \item[\(E_{ij}\)] Optional edge features between node \(i\) and node \(j\) (for example, relative distance or IoU).
    \item[\(M_{ij}\)] The message from node \(i\) to node \(j\), computed as
    \[
    M_{ij} = \mathrm{Conv}_{1\times1}\Bigl(\mathrm{Concat}(X_i,\,X_j,\,E_{ij})\Bigr).
    \]
    \item[\(\mathcal{A}\)] The deep CNN aggregator that refines aggregated messages using \(3\times3\) convolutions, batch normalization, dropout, and ReLU.
    \item[\(m_j\)] The aggregated message at node \(j\), defined by
    \[
    m_j = \sum_{i \in \mathcal{N}(j)} M_{ij}.
    \]
    \item[\(X'_j\)] The updated node feature after a residual update, given by
    \[
    X'_j = X_j + \mathcal{A}(m_j).
    \]
    \item[\(z_j\)] The vector obtained by spatially pooling the feature map \(X'_j\) (typically via average pooling).
    \item[\(\hat{y}_j\)] The class logits for node \(j\), computed by a linear classifier as
    \[
    \hat{y}_j = W\,z_j + b.
    \]
    \item[\(\theta_{cnn}\)] The set of parameters for the CNN encoder.
    \item[\(\alpha,\ \beta\)] Coefficients for weighting different components in the composite loss function.
    \item[\(\gamma\)] The coefficient controlling the influence of the AUC-based ranking loss component.
    \item[\(\hat{s}_{i,y_i}\) and \(\hat{s}_{i,j}\)] The predicted score for the true class of sample \(i\) and the predicted score for an incorrect class \(j\), respectively, used in the AUC-based ranking loss.
    \item[\(C\)] The number of classes for the classification task.
    \item[\(\mathrm{IoU}_{\mathrm{gt}}\)] The ground-truth intersection-over-union value used in the composite loss.
\end{description}

\section*{Appendix B: Detailed Architecture and Hyperparameters}
\addcontentsline{toc}{subsection}{Appendix B: Detailed Architecture and Hyperparameters}

Key components of the TGraphX architecture include:

\begin{itemize}
    \item \textbf{Patch Extraction:}  
    \begin{itemize}
        \item The input image \(I \in \mathbb{R}^{H \times W \times C_{in}}\) is divided into \(N\) patches \(P_i\) of size \(patch\_H \times patch\_W\).
    \end{itemize}
    \item \textbf{Pre-Encoder (Optional):}  
    \begin{itemize}
        \item An optional module (or a pretrained ResNet block) processes each patch, yielding \(F_i \in \mathbb{R}^{H' \times W' \times C_{pre}}\).
    \end{itemize}
    \item \textbf{CNN Encoder:}  
    \begin{itemize}
        \item Composed of \(L\) convolutional blocks with \(3 \times 3\) kernels, each followed by batch normalization, ReLU, and dropout (e.g., with dropout rate 0.3).
        \item Residual connections are used to ease training.
        \item Outputs spatial feature maps \(X_i \in \mathbb{R}^{C_{\mathrm{cnn}} \times H_2 \times W_2}\).
    \end{itemize}
    \item \textbf{Graph Construction:}  
    \begin{itemize}
        \item \textbf{Nodes:} Each feature map \(X_i\) forms a node, preserving its three-dimensional structure.
        \item \textbf{Edges:} Edges are defined based on spatial (or semantic) proximity; specifically, an edge is formed between nodes \(i\) and \(j\) if \(d(P_i,P_j) \leq \tau\).
        \item Batching is handled by merging graphs into a \texttt{GraphBatch} that properly offsets node indices.
    \end{itemize}
    \item \textbf{Message Passing and Aggregation:}  
    \begin{itemize}
        \item Messages \(M_{ij}\) are computed using a \(1 \times 1\) convolution on the concatenation of \(X_i\), \(X_j\), and optionally \(E_{ij}\).
        \item Aggregated messages \(m_j\) (over the neighborhood \(\mathcal{N}(j)\)) are refined by a deep CNN aggregator \(\mathcal{A}\), and a residual update \(X'_j = X_j + \mathcal{A}(m_j)\) is applied.
    \end{itemize}
    \item \textbf{Spatial Pooling and Classification:}  
    \begin{itemize}
        \item Average pooling is applied over the spatial dimensions of \(X'_j\) to produce a vector \(z_j\).
        \item A linear classifier maps \(z_j\) to the final class logits \(\hat{y}_j\).
    \end{itemize}
    \item \textbf{Loss Function:}  
    \begin{itemize}
        \item A composite loss function is used, for example,
        \[
        \mathcal{L} = \alpha\,\mathcal{L}_{\mathrm{CE}}(\hat{y}, y) + \beta\,\Bigl\lVert \widehat{\mathrm{IoU}} - \mathrm{IoU}_{\mathrm{gt}}\Bigr\rVert^2,
        \]
        where \(\mathcal{L}_{\mathrm{CE}}\) is the cross-entropy loss and the second term penalizes discrepancies in bounding box alignment.
    \end{itemize}
\end{itemize}

\section*{Appendix C: Pseudocode for the TGraphX Pipeline}
\addcontentsline{toc}{subsection}{Appendix C: Pseudocode for the TGraphX Pipeline}

The pseudocode below outlines the main steps in the TGraphX pipeline. All mathematical symbols are represented using standard LaTeX commands.

\begin{verbatim}
Input: Image I, where I in R^(H x W x C_in)
Output: Final classification output

1. // Patch Extraction
   Divide I into N patches {P_i} of size (patch_H x patch_W)

2. // Optional Pre-Encoder
   For each patch P_i:
       If use_preencoder is true:
           Set F_i = PreEnc(P_i)
       Else:
           Set F_i = P_i

3. // CNN Encoder
   For each feature map F_i:
       Compute X_i = CNNEnc(F_i; theta_cnn)
       // X_i has shape: [C_cnn, H2, W2]

4. // Graph Construction
   Construct graph G = (V, E), where:
       V = {X_i}  (each X_i is a node)
       E = {(i, j) such that d(P_i, P_j) <= tau}
       Optionally, attach edge features E_ij

5. // Message Passing
   For each edge (i, j) in E:
       Compute M_ij = Conv_1x1(Concat(X_i, X_j, E_ij))
   For each node j:
       Set m_j = sum over i in N(j) of M_ij

6. // Aggregation and Residual Update
   For each node j:
       Compute X_prime_j = X_j + Aggregator(m_j)
       // Aggregator applies multiple 3x3 convolutions, BN, dropout, and ReLU

7. // Spatial Pooling and Classification
   For each node j:
       Compute z_j = AveragePool(X_prime_j)
       Compute y_hat_j = Linear(z_j)
       // y_hat_j are the class logits for node j

8. // Loss Computation (if training)
   Compute composite loss:
       L = alpha * CE(y_hat, y) + beta * || IoU_pred - IoU_gt ||^2

Return: Final classification output (aggregated y_hat values)
\end{verbatim}

\section*{Appendix D: Additional Implementation Notes}
\addcontentsline{toc}{subsection}{Appendix D: Additional Implementation Notes}

\begin{itemize}
    \item \textbf{Graph Batching:}  
          Multiple \texttt{Graph} objects are merged into a \texttt{GraphBatch} for efficient parallel processing. Node indices are offset to prevent collisions.
    \item \textbf{Training Details:}  
          TGraphX is implemented in PyTorch. Standard optimization techniques (learning rate scheduling, weight decay, and mini-batch gradient descent) are used. Hyperparameters (e.g., learning rate, batch size, dropout rate) are tuned per dataset.
    \item \textbf{Modularity:}  
          The design permits easy swapping of components (such as alternative pre-encoders or message passing methods) and supports extensions to multimodal data.
    \item \textbf{Residual Connections:}  
          Residual connections are employed in both the CNN encoder and the deep CNN aggregator to ensure stable gradient flow in deep architectures.
\end{itemize}

\newpage
\renewcommand{\bibname}{}  
\renewcommand{\refname}{}  
\section{References}
\bibliographystyle{unsrt}
\bibliography{references}

\end{document}